\documentclass{article}

\usepackage{arxiv}

\usepackage[utf8]{inputenc} % allow utf-8 input
\usepackage[T1]{fontenc}    % use 8-bit T1 fonts
\usepackage{hyperref}       % hyperlinks
\usepackage{url}            % simple URL typesetting
\usepackage{booktabs}       % professional-quality tables
\usepackage{amsfonts}       % blackboard math symbols
\usepackage{nicefrac}       % compact symbols for 1/2, etc.
\usepackage{microtype}      % microtypography
\usepackage{lipsum}		% Can be removed after putting your text content
\usepackage{graphicx}
\usepackage[numbers]{natbib}
\usepackage{doi}
\usepackage{amsmath}
\usepackage{cleveref}
\usepackage{todonotes}
\usepackage{subcaption}
\usepackage{lscape} 
\usepackage[pagewise]{lineno}%\linenumbers

\title{Multifidelity Kolmogorov-Arnold Networks}

\date{} 					% Or removing it

\author{
        \href{https://orcid.org/0000-0002-6411-6198}{\includegraphics[scale=0.06]{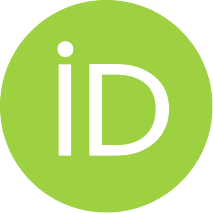}\hspace{1mm}Amanda A. Howard}\\
	Pacific Northwest National Laboratory\\
	Richland, WA 99354 \\
	\texttt{amanda.howard@pnnl.gov} \\
    \And  
        \href{https://orcid.org/0009-0001-5361-3105}{\includegraphics[scale=0.06]{orcid.pdf}\hspace{1mm}Bruno Jacob}\\
	Pacific Northwest National Laboratory\\
	Richland, WA 99354 \\
	\texttt{bruno.jacob@pnnl.gov} \\
    \And
        \href{https://orcid.org/0000-0002-9928-5637}{\includegraphics[scale=0.06]{orcid.pdf}\hspace{1mm}Panos Stinis} \\
	Pacific Northwest National Laboratory\\
	Richland, WA 99354 \\
        \texttt{panagiotis.stinis@pnnl.gov}  
}

\renewcommand{\shorttitle}{Multifidelity KANs}

\hypersetup{
}

\begin{document}
\maketitle

\begin{abstract}
We develop a method for multifidelity Kolmogorov-Arnold networks (KANs), which use a low-fidelity model along with a small amount of high-fidelity data to train a model for the high-fidelity data accurately. Multifidelity KANs (MFKANs) reduce the amount of expensive high-fidelity data needed to accurately train a KAN by exploiting the correlations between the low- and high-fidelity data to give accurate and robust predictions in the absence of a large high-fidelity dataset. In addition, we show that multifidelity KANs can be used to increase the accuracy of physics-informed KANs (PIKANs), without the use of training data. 
\end{abstract}

% keywords can be removed

\begin{figure}[h]
    \centering
    \includegraphics[width=0.7\textwidth]{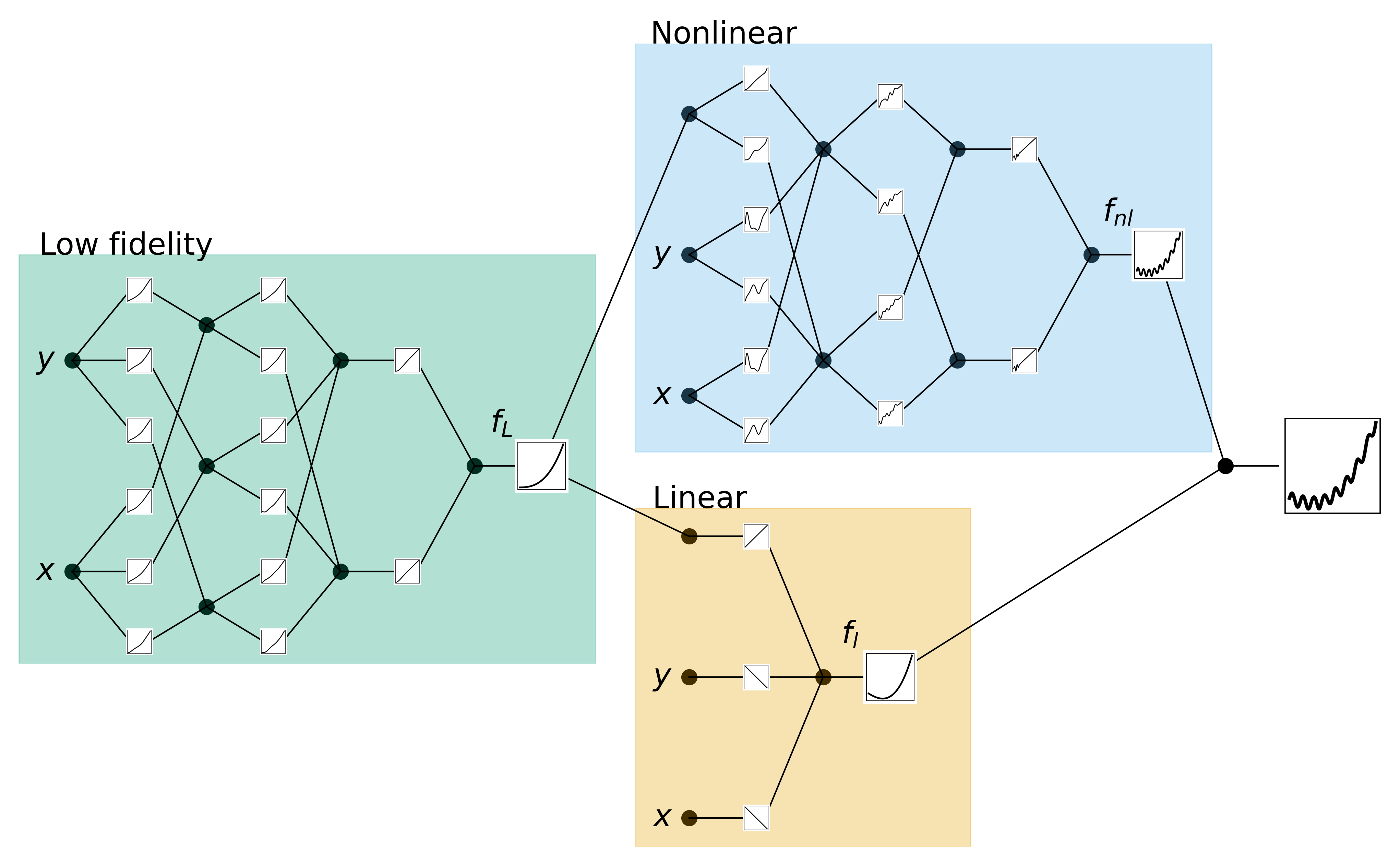}
    \caption*{\emph{Graphical abstract:} Multifidelity KANs accurately learn from low-fidelity and high-fidelity data simultaneously.}
    \label{fig:graph_abs}
\end{figure}

\section{Introduction}

In recent years, scientific machine learning (SciML) has emerged as a paradigm for modeling physical systems \cite{karniadakis2021physics, baker2019workshop, carter2023advanced}. Typically using the theory of multilayer perceptrons (MLPs), SciML has shown great success in modeling a wide range of applications, however, data-informed training struggles when high-quality data is not available. 

Kolmogorov-Arnold networks  (KANs) have recently been developed as an alternative to MLPs \cite{liu2024kan, liu2024kan20}. KANs use the Kolmogorov-Arnold Theorem as inspiration and can offer advantages over MLPs in some cases, such as for discovering interpretable models. However, KANs have been shown to struggle to reach the accuracy of MLPs, particularly without modifications \cite{yu2024kan, zeng2024kan, shen2024reduced, shukla2024comprehensive}. In the short time since the publication of \cite{liu2024kan}, many variations of KANs have been developed, including physics-informed KANs (PIKANs)\cite{shukla2024comprehensive}, KAN-informed neural networks (KINNs)\cite{wang2024kolmogorov}, temporal KANs \cite{genet2024tkan}, wavelet KANs \cite{bozorgasl2024wav}, graph KANs \cite{kiamari2024gkan, decarlo2024kolmogorovarnoldgraphneuralnetworks, bresson2024kagnnskolmogorovarnoldnetworksmeet}, Chebyshev KANs (cKANs) \cite{ss2024chebyshev}, convolutional KANs \cite{bodner2024convolutional}, ReLU-KANs \cite{qiu2024relu}, Higher-order-ReLU-KANs (HRKANs) \cite{so2024higher}, fractional KANs \cite{aghaei2024fkan}, finite basis KANs \cite{howard2024finite}, deep operator KANs \cite{abueidda2024deepokan}, and others. KANs have been applied to fluid dynamics \cite{toscano2024inferring, kashefi2024kolmogorov}, time-series analysis \cite{vaca2024kolmogorov}, satellite image classification \cite{cheon2024kolmogorov}, abnormality detection \cite{huang2024abnormality}, and computer vision \cite{azam2024suitability, cheon2024demonstrating}, among other applications \cite{nagai2024kolmogorov}. 

Multifidelity machine learning uses two or more datasets to train a network more accurately than using either dataset alone \cite{fernandez2016review}. The most common framework is bi-fidelity training with two datasets, denoted by the low-fidelity and high-fidelity datasets. For scientific applications, low-fidelity data is often data generated by a lower-order model or on a coarser mesh. In this way, the low-fidelity data is less expensive to generate, so it is possible to sample a large low-fidelity dataset. However, it is less accurate. The high-fidelity dataset is a small dataset of highly accurate data. The goal of multifidelity training is to train simultaneously with low- and high-fidelity data. Multifidelity machine learning is an increasingly important research field due to the cost of generating and storing highly accurate training data \cite{fernandez2016review, peherstorfer2018survey, penwarden2022multifidelity}. Several approaches for multifidelity training have been developed, including transfer learning \cite{song2022transfer, chakraborty2021transfer, propp2024transferlearningmultidimensionaldata, lejeune2021exploring, li2022line, de2022neural, jiang2023use, de2020transfer, aliakbari2022predicting}, intermediate approaches \cite{guo2022multi, chen2023feature}, and composite multifidelity neural networks \cite{meng2020composite}. Composite multifidelity neural networks have been particularly successful, and have been extended to a large number of variations, including PINNs \cite{meng2020composite, howard2023stacked}, continual learning \cite{howard2023continual}, deep operator networks \cite{lu2022multifidelity, howard2022onets, de2023bi},  Bayesian neural networks \cite{meng2021multi}, and a large number of applications \cite{su2023multifidelity, ramezankhani2022data}, \textit{e.g.,} aerodynamics \cite{zhang2021multi, he2020multi}, material properties \cite{islam2021extraction}, rotor dynamics \cite{khamari2023novel}, and rheology \cite{chiniforooshan2023data, boodaghidizaji2022multi, saadat2024data, mahmoudabadbozchelou2021data}.

While KANs have been widely applied, training KANs still relies on having sufficiently abundant, accurate data and KANs (like MLPs) can struggle to train in the presence of noisy data \cite{zeng2024kan, shen2024reduced}. Also, compared to MLPs, KANs can struggle to train with sparse data \cite{pourkamali2024kolmogorov, poeta2024benchmarking}. In this work, we introduce \textit{multifidelity KANs} (MFKANs) which allow for training simultaneously with two datasets. MFKANs reduce the need for a large high-fidelity dataset by incorporating training with a second, low-fidelity dataset. We show that MFKANs can accurately learn linear and nonlinear correlations between the low- and high-fidelity datasets. In addition, MFKANs can be extended to physics-informed KANs \cite{shukla2024comprehensive}, when no high-fidelity or low-fidelity data is available, but the physics governing the system is known. 

\section{Method} \label{sec:method}
\subsection{KANs}
Kolmogorov-Arnold networks \cite{liu2024kan} approximate a multivariate function $f(\mathbf{x})$ by a model of the form
\begin{equation}
    f(\mathbf{x}) \approx \sum_{i_{L-1}=1}^{n_{L-1}}\varphi_{L-1, i_{L}, i_{L-1}}
    \left(\sum_{i_{L-2}=1}^{n_{L-2}} \ldots
    \left(\sum_{i_{2}=1}^{n_{2}}\varphi_{2, i_3, i_2}
    \left(\sum_{i_{1}=1}^{n_{1}}\varphi_{1, i_2, i_1}
    \left(\sum_{i_{0}=1}^{n_{0}}\varphi_{0, i_1, i_0}(x_{i_0})
    \right)
    \right)
    \right) \ldots
    \right), \label{eq:KAN}
\end{equation}
inspired by the Kolmogorov-Arnold Theorem. $L$ is the number of layers in the KAN, $\{n_j\}_{j=0}^{L}$ is the number of nodes per layer, and $\phi_{i,j, k}$ are the univariate activation functions. We denote the right-hand side of~\cref{eq:KAN} as $\mathcal{K}(x)$. The activation functions are represented (on a grid with $g$ points) by a weighted combination of a basis function $b(x)$ and a B-spline, 
\begin{equation}
    \phi(x) = w_b b(x) + w_s \text{spline}(x) \label{eq:phi_format},
\end{equation}
where 
$$
    b(x) = \frac{x}{1+e^{-x}},
$$
and
$$
    \text{spline}(x) = \sum_i c_i B_i(x).
$$
Here, $B_i(x)$ is a polynomial of degree $k$. $c_i, w_b,$ and $w_s$ are trainable parameters if not predetermined by the user. 

KANs evaluate the B-splines on a precomputed grid, which should align with the domain of the data. For example, in one dimension a domain $[a, b]$  with a grid with $g_1$ intervals has grid points $\{t_0 = a, t_1, t_2, \ldots, t_{g_1}=b\}$; cf.~\cite{liu2024kan}. In grid extension \cite{liu2024kan, rigas2024adaptive} a new, fine-grained spline is fitted to a coarse-grained spline, increasing the expressivity of the KAN. The coarse splines are transferred to the fine splines following the procedure in \cite{rigas2024adaptive}. The method in which the grid is selected has a significant impact on the training \cite{rigas2024adaptive}, although we leave a careful exploration of this topic for future work. 

Single-fidelity KANs are trained to minimize the mean squared error (MSE) between the data and the prediction. For data-informed training we consider a dataset of labeled pairs $\left\{ (x_i, f(x_i))\right\}_{i=1}^N$ and minimize the loss function
\begin{equation}
  \mathcal{L}(\theta) = \frac{1}{N}\sum_{i=1}^{N}\left[ \mathcal{K}(x_i; \theta)-f(x_i)\right]^2.
\end{equation}
Here, $\theta$ is the set of trainable parameters. Physics-informed KANs (PIKANs) \cite{shukla2024comprehensive} have emerged as an alternative to physics-informed neural networks (PINNs) \cite{raissi2019physics, karniadakis2021physics, cai2021physics}, where a physics-informed loss can be implemented in a manner analogous to PINNs \cite{raissi2019physics}. We refer to recent work for a summary of PIKANs \cite{shukla2024comprehensive}, and note that there are many recent extensions and variants \cite{rigas2024adaptive, ranasinghe2024ginn, so2024higher, patra2024physics, wang2024kolmogorov, shuai2024physics, yu2024sinc}. 

\subsection{Multifidelity KANs}
In this work, we take inspiration from composite multifidelity neural networks \cite{meng2020composite}, which consist of three MLPs. The first MLP learns a model of the low-fidelity training data. The second and third MLPs are added together to learn the correlation between the output of the low-fidelity model and the high-fidelity data, to provide an accurate prediction of the high-fidelity data. In particular, the second MLP learns the linear correlation between the low-fidelity model and the high-fidelity data, and the third MLP learns the nonlinear correlation. Ideally, the low-fidelity and high-fidelity datasets are highly correlated, so the linear correlation accounts for the bulk of the correlation, while the nonlinear correlation is a small correction. This structure allows for the nonlinear network to be smaller, increasing the robustness of training when only a small amount of high-fidelity data is available.  

The foundations of multifidelity neural networks as presented in \cite{meng2020composite} translate well to multifidelity KANs (MFKANs). An MFKAN consists of three blocks: a low-fidelity KAN ($\mathcal{K}_{L}$), a linear KAN ($\mathcal{K}_{l}$), and a nonlinear KAN ($\mathcal{K}_{nl}$). The low-fidelity KAN learns a surrogate for the low-fidelity data or physics. Then, the output from the low-fidelity KAN is appended to the input parameters and passed as inputs to the linear and nonlinear KANs. The linear KAN learns the linear correlation between the input variables, the output of the low-fidelity KAN, and the high-fidelity data. The nonlinear KAN learns a nonlinear correction to the linear correlation, to account for general datasets which may not be strictly linearly correlated. An example of this structure is shown in Fig. \ref{fig:MF_KAN}.

We consider the case where we have two datasets, a low-fidelity dataset consisting of labeled pairs $\left\{ (x_i, f_{L}(x_i))\right\}_{i=1}^{N_{LF}}$ and a high-fidelity dataset 
$\left\{ (x_j, f_{H}(x_j))\right\}_{j=1}^{N_{HF}}$. Note that there is no requirement that $\{x_j\}_{j=1}^{N_{HF}}$ is a subset of $\{x_i\}_{i=1}^{N_{LF}}$. In the following sections, we provide some details on the specifics of each block.

\subsubsection{Low-fidelity block} In general, the low-fidelity block can be a low-fidelity numerical model implemented directly, a standard KAN, or an MLP or other machine learning model. For the purposes of this work, we only consider KANs for the low-fidelity block except in Sec. \ref{sec:Test7}, where we show an example of using a low-fidelity numerical model directly. The low-fidelity KAN is pretrained (as a single-fidelity KAN) using the low-fidelity data. The weights of the low-fidelity KAN are then frozen while training the high-fidelity networks. While training all networks simultaneously is possible, we found that it greatly increased the computational time. 

The low-fidelity loss function with trainable parameters $\theta_{L}$ is given by
\begin{equation}
  \mathcal{L}_{L}(\theta_{L}) = \frac{1}{N_{LF}}\sum_{i=1}^{N_{LF}}\left[ \mathcal{K}_{L}(x_i; \theta_{L})-f_{L}(x_i)\right]^2.
\end{equation}

\subsubsection{High-fidelity block} 
The high-fidelity prediction is a convex combination of the linear and nonlinear networks, given by
\begin{equation}
    \mathcal{K}_{H}(\mathbf{x}) = \alpha \mathcal{K}_{nl}(\mathbf{x}) + (1-\alpha)\mathcal{K}_{l}(\mathbf{x}),
\end{equation}
where $\alpha$ is a trainable parameter. 
The high-fidelity loss function is modified to 
\begin{equation}
  \mathcal{L}_{H}(\theta_{H}, \alpha) = \frac{1}{N_{HF}}\sum_{j=1}^{N_{HF}}\left[ \mathcal{K}_{H}(x_j; \theta_{H})-f_{H}(x_j)\right]^2 + \lambda_\alpha \alpha^n + w \sum_{l=0}^{L-1}||\Phi_{nl}||, \label{eq:HF_Loss}
\end{equation}
where $\theta_{H} = \{\theta_{nl}, \theta_{l}\}$ is the set of trainable network parameters and $n$ is a scalar, typically taken as $n=4$. This minimizes $\alpha$, which forces the method to learn the maximum linear correlation. If $n$ is too small, forcing $\alpha$ to be smaller, the method may not learn the full nonlinear correlation, which reduces the expressivity of the method. If a strong nonlinear correlation is expected between the high-fidelity and low-fidelity data, it is possible to choose a larger value of $n$, although this training may require more high-fidelity data. $\lambda_\alpha$ is a parameter chosen before training. The set of trainable parameters $\theta_{HF}$ includes all trainable parameters of both the linear and nonlinear KANs. The final term in the loss function Eq. \cref{eq:HF_Loss} is the sum of the mean squared value of each KAN layer in the nonlinear network, inspired by \cite{liu2024kan} and given by
\begin{equation}
    ||\Phi_{nl}|| = \frac{1}{n_{in}n_{out}}\sum_{i=1}^{n_{in}}\sum_{j=1}^{n_{out}}|\phi_{i,j}^{nl}|^2. 
\end{equation}
While \cite{liu2024kan} minimizes the $L_1$ norm to learn sparse solutions, here we use the minimization to prevent overfitting to possibly sparse high-fidelity data. We typically take $w=0$ or $w=1$, depending on any \textit{a priori} known correlation between the low-fidelity and high-fidelity data.

\begin{figure}[h]
    \centering    \includegraphics[width=0.7\textwidth]{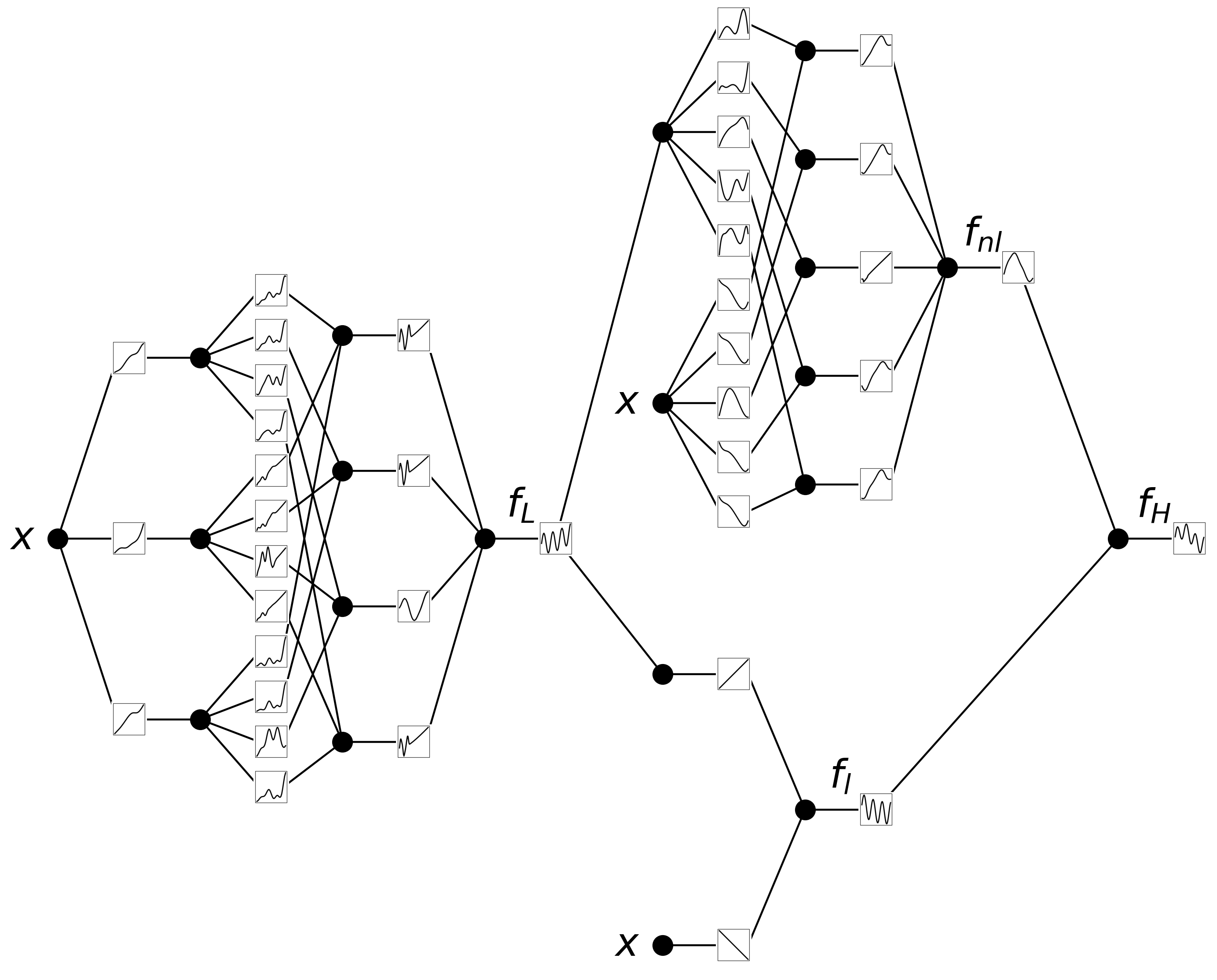}
    \caption{Example multifidelity KAN. The low-fidelity data is given by $f_L(x) = x^2+ \sin(20x)$ and $f_H(x) = f_L(x)-x+\sin(5x)$ for $x\in[0, 1]$. It is evident that the linear KAN learns the linear correlations and the nonlinear KAN outputs the nonlinear correlation ($\sin(5x)$).}
    \label{fig:MF_KAN}
\end{figure}

\subsubsection{Linear block} A linear correlation is a first-degree polynomial, which makes this block easy to implement as a KAN by taking the polynomial degree of the basis as $k=1$. We also must take $w_b = 0$ and $w_s = 1$ in Eq. \ref{eq:phi_format}. We use only two grid points (\textit{i.e.,} $g_1 = 1$ so $t_0 = a$ and $t_1 = b$). This allows learning a global linear correlation that is constant across the full domain. For all examples in this work, the linear block does not contain a hidden layer. 

KANs do offer additional flexibility over MLPs, though. If it is known that the low-fidelity data and the high-fidelity data are linearly correlated, but with different correlations in different parts of the domain, we could increase the number of grid points to learn additional linear correlations. Ideally, the grid points should correspond with the boundaries of the domains of each of the known linear correlations. We leave an exploration of this idea for future work.

\subsubsection{Nonlinear block} The nonlinear block is a standard KAN of degree $k > 1$. We still take $w_s = 1$ in Eq. \ref{eq:phi_format}, as we find this improves the training ($w_b$ is a trainable parameter.)

\subsubsection{Implementation} For the examples listed in this paper, the code is implemented in Jax \cite{jax2018github}, using the jaxKAN package \cite{Rigas_jaxKAN_A_JAX-based_2024, rigas2024adaptive}.

To report errors we consider the relative $\ell_2$ error, given by
\begin{equation}
  \frac{||\mathcal{K}_{i}(x; \theta_{i})-f_{j}(x)||_2}
  {||f_{j}(x)||_2}
\end{equation}
where $i$ and $j$ denote either low-fidelity or high-fidelity.

\section{Results}
\subsection{Test 1: Jump function with a linear correlation} \label{sec:Test1}
\begin{figure}[h]
    \centering
    \includegraphics[width=0.9\textwidth]{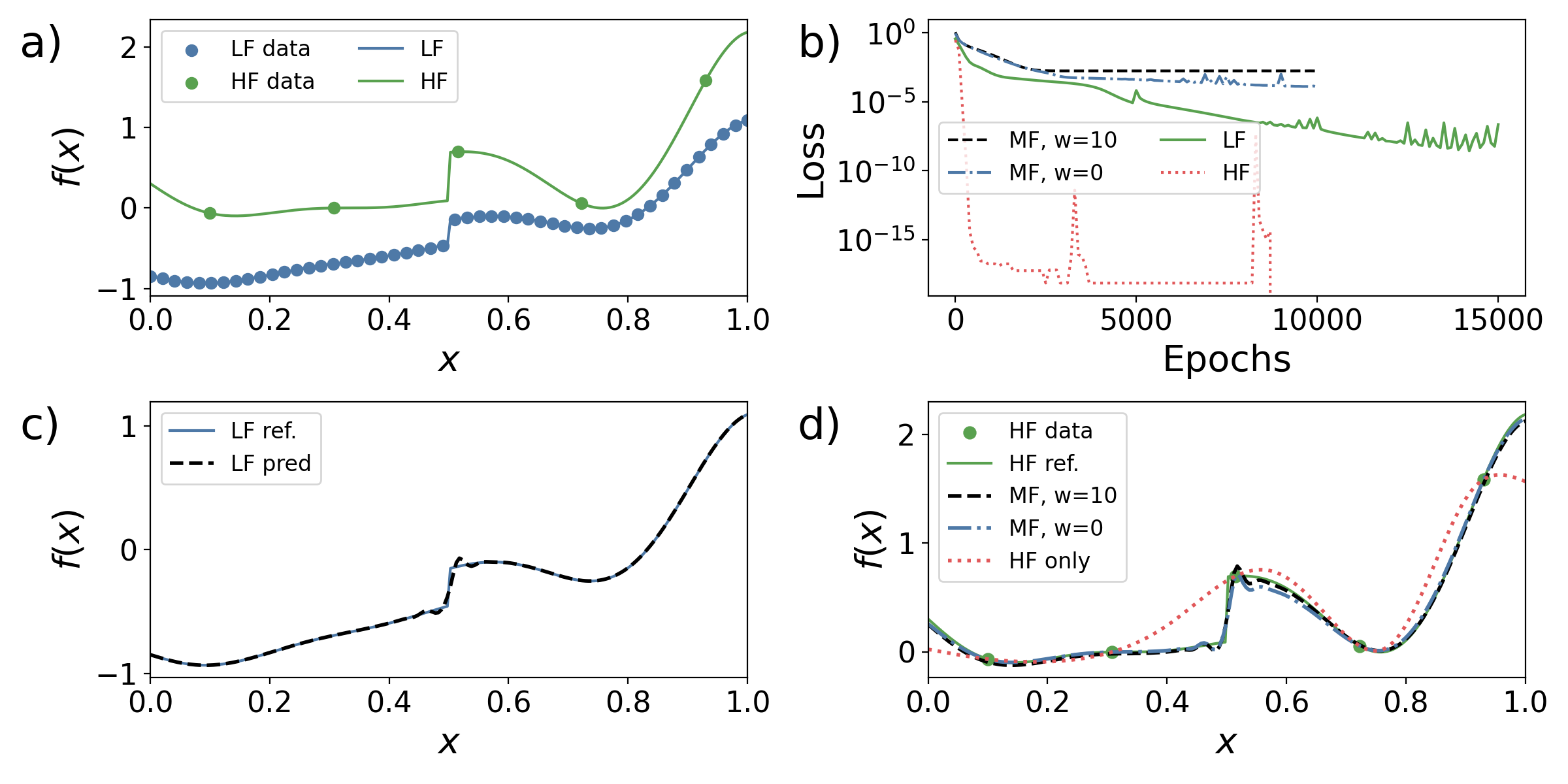}
    \caption{Results for Test 1 with $N_{LF} = 50$. a) Plotted low- and high-fidelity data points and functions for reference. b) Loss values for multifidelity, low-fidelity, and high-fidelity training. c) Low-fidelity reference function and low-fidelity prediction. d) High-fidelity reference function and multifidelity and high-fidelity predictions. }
    \label{fig:Test1a}
\end{figure}

\begin{figure}[h]
    \centering
    \includegraphics[width=0.9\textwidth]{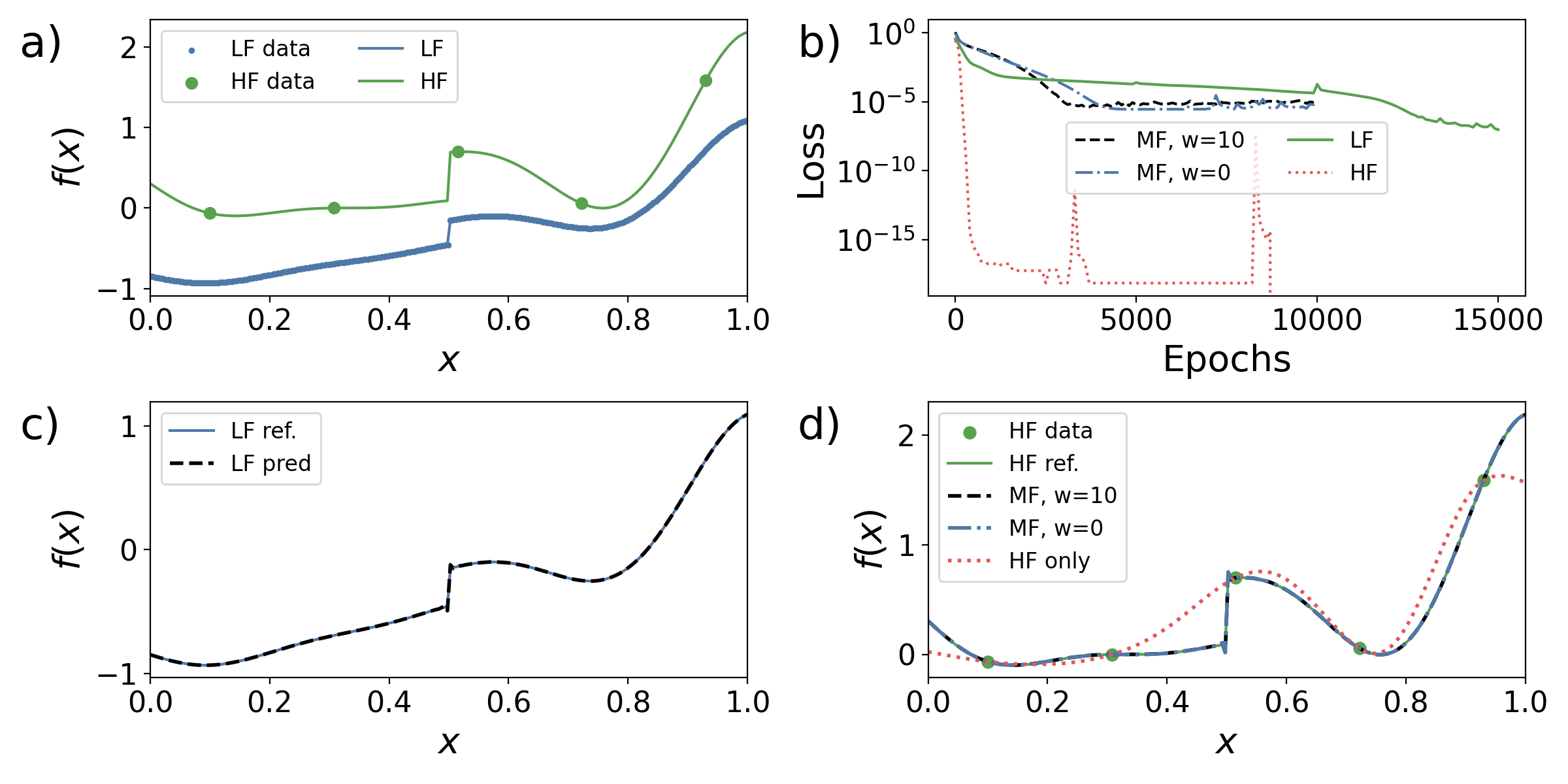}
    \caption{Results for Test 1 with $N_{LF} = 300$. a) Plotted low- and high-fidelity data points and functions for reference. b) Loss values for multifidelity, low-fidelity, and high-fidelity training. c) Low-fidelity reference function and low-fidelity prediction. d) High-fidelity reference function and multifidelity and high-fidelity predictions. }
    \label{fig:Test1b}
\end{figure}

We begin by considering a jump function with a linear correlation. In this example, we have sparse high-fidelity data, which is not sufficient to capture the jump. 
\begin{align}
    f_L(x) &= \begin{cases} 
      0.1\left[0.5(6x-2)^2 \sin(12x-4) + 10(x-0.5)-5\right] & x\leq 0.5, \\
      0.1\left[0.5(6x-2)^2 \sin(12x-4) + 10(x-0.5)-2\right] & x > 0.5,  
   \end{cases}\\
f_H(x) &= 2f_L(x) -2x + 2,
\end{align}
for $x \in [0, 1].$ We take $N_{LF}= 50$ low-fidelity data points evenly distributed in $[0,1]$ and $N_{HF} =5$ high-fidelity data points evenly spaced in [0.1, 0.93]. Results are shown in Fig. \ref{fig:Test1a} and Fig. \ref{fig:Test1b}. 

We first consider training only on high-fidelity data. Although the network trains to a very small MSE on the high-fidelity training data ($<10^{-15}$) (Fig. \ref{fig:Test1a}b), the resulting prediction has a large relative $\ell_2$ error of 0.307 because of the sparsity of the high-fidelity data. We then consider the multifidelity training with $w=0$ and $w=1$. One thing to note is that the low-fidelity prediction, even with the large amount of low-fidelity data, struggles to capture the jump accurately. This could be due to the B-splines used in the KANs basis having to interpolate the jump. The relative $\ell_2$ error of the low-fidelity prediction is 0.0231. The high-fidelity predictions have very similar relative $\ell_2$ errors, of $0.0651$ and $0.0571$ for $w=0$ and $w=10$, respectively. Even with sparse high-fidelity data, the multifidelity prediction is able to accurately capture the jump. 

To better understand how the error in the low-fidelity predictions due to having discrete data is propagated through to the multifidelity predictions, we consider a separate test with  $N_{LF}= 300$, shown in Fig. \ref{fig:Test1b}. With the additional low-fidelity data, the relative $\ell_2$ error of the low-fidelity prediction drops to 0.0054, approximately a factor of four. The multifidelity predictions drop by a corresponding amount, to $0.0105$ for $w=0$ and $0.0102$ for $w=10$.

\subsection{Test 2: Nonlinear correlation}\label{sec:Test2}
\begin{figure}[h]
    \centering
    \includegraphics[width=0.9\textwidth]{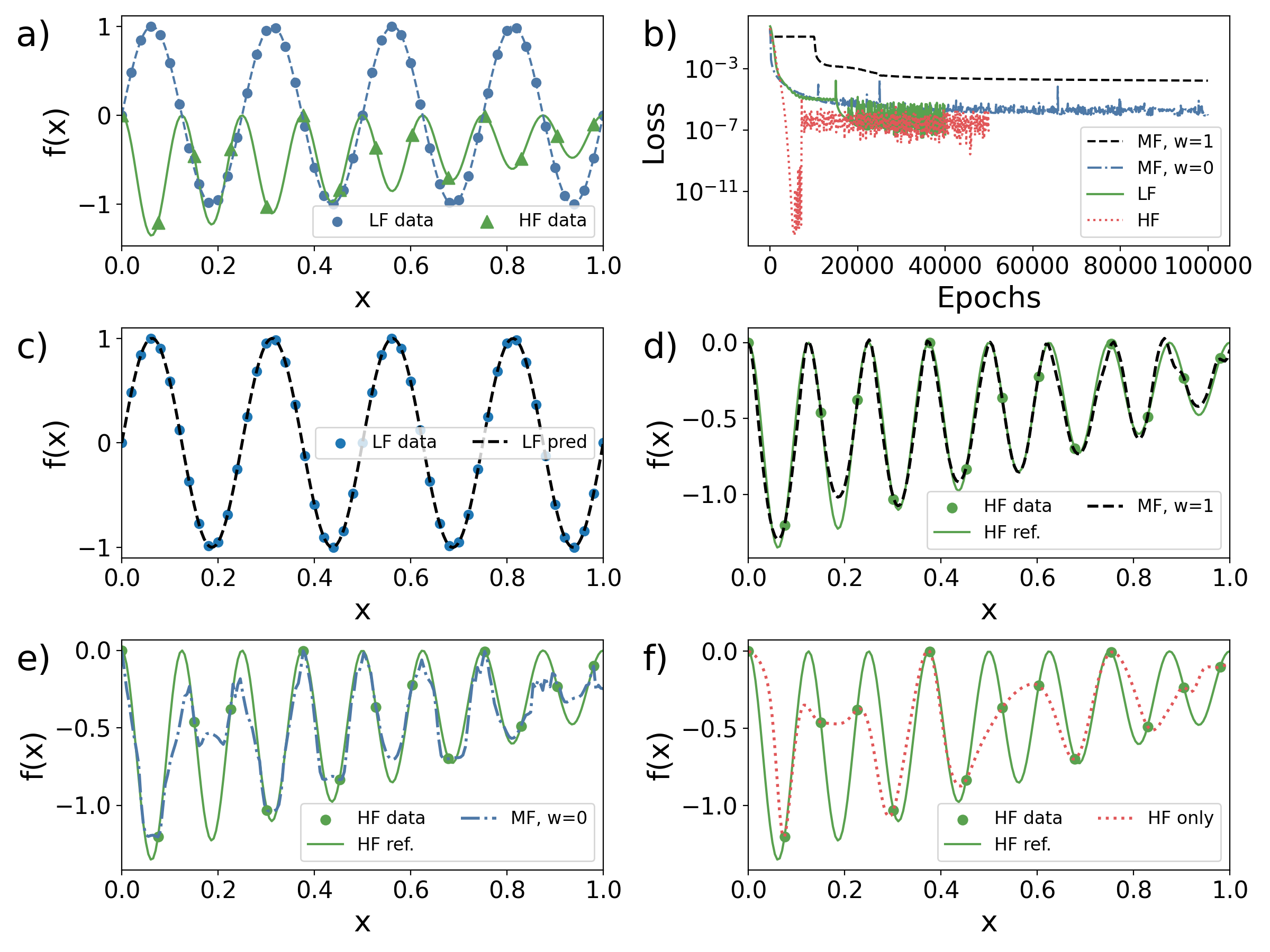}
    \caption{Results for Test 2. a) Reference solutions and training data. b) Loss curves for single-fidelity and multifidelity training. c) Low-fidelity reference solution and prediction. d) High-fidelity reference solution and multifidelity prediction with $w=1.$ e) High-fidelity reference solution and multifidelity prediction with $w=0.$, f) High-fidelity reference solution and high-fidelity prediction. }
    \label{fig:Test2}
\end{figure}

We next test a nonlinear correlation between the low-fidelity and high-fidelity data, taken from \cite{meng2020composite}. The equations for the low- and high-fidelity data are given by:
\begin{align}
    f_L(x) &= \sin(8 \pi x),\\
    f_H(x) &= (x-\sqrt{2}) \sin^2(8 \pi x),
\end{align}
We take $N_{LF} = 51$ and $N_{HF} = 14$ on $[0, 1]$. Results are shown in Fig. \ref{fig:Test2}. When $w=0$, the nonlinear correlation suffers from overfitting the sparse high-fidelity data, which is shown in Fig. \ref{fig:Test2}e. When $w=1$, the overfitting is reduced. Clearly, training with only the high-fidelity data in Fig. \ref{fig:Test2}f does not  accurately capture the high-fidelity reference. 

\subsection{Test 3: two-dimensional nonlinear correlation}\label{sec:Test3}
\begin{figure}[ht]
    \centering
    \includegraphics[width=0.7\textwidth]{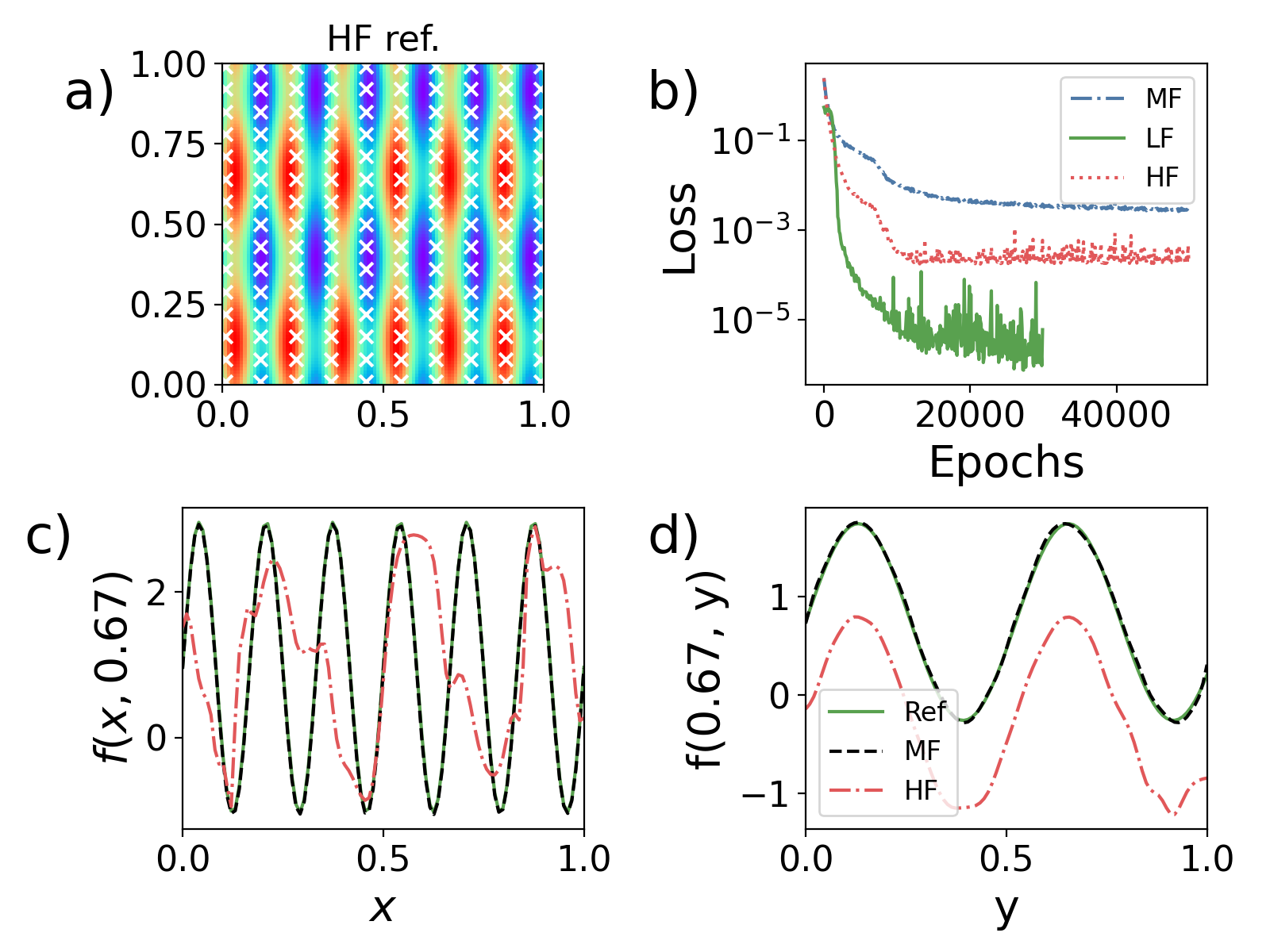}
    \caption{Results for Test 3. a) Depiction of the high-fidelity data locations. b) Loss curves for low-fidelity, high-fidelity, and multifidelity training. c) Trained predictions along the line $y = 0.67$. d) Trained predictions along the line $x = 0.67$.}
    \label{fig:Test3a}
\end{figure}

To further test the method, we turn to a nonlinear correlation in two dimensions, given by
\begin{align}
    f_L(x, y) &= \sin(12 \pi x), \\
    f_H(x, y) &= 2f_L(x, y) + \sin(12y),
\end{align}
for $x, y \in [0, 1] \times [0, 1].$
The low-fidelity data is selected on a 100$\times$100 mesh, so $N_{LF} = 10^4$. The high-fidelity data is selected on a $10 \times 15$ mesh, with $N_{HF} = 150$. The high-fidelity data locations are shown in Fig. \ref{fig:Test3a}a. 

\begin{figure}[h]
    \centering
    \includegraphics[width=0.9\textwidth]{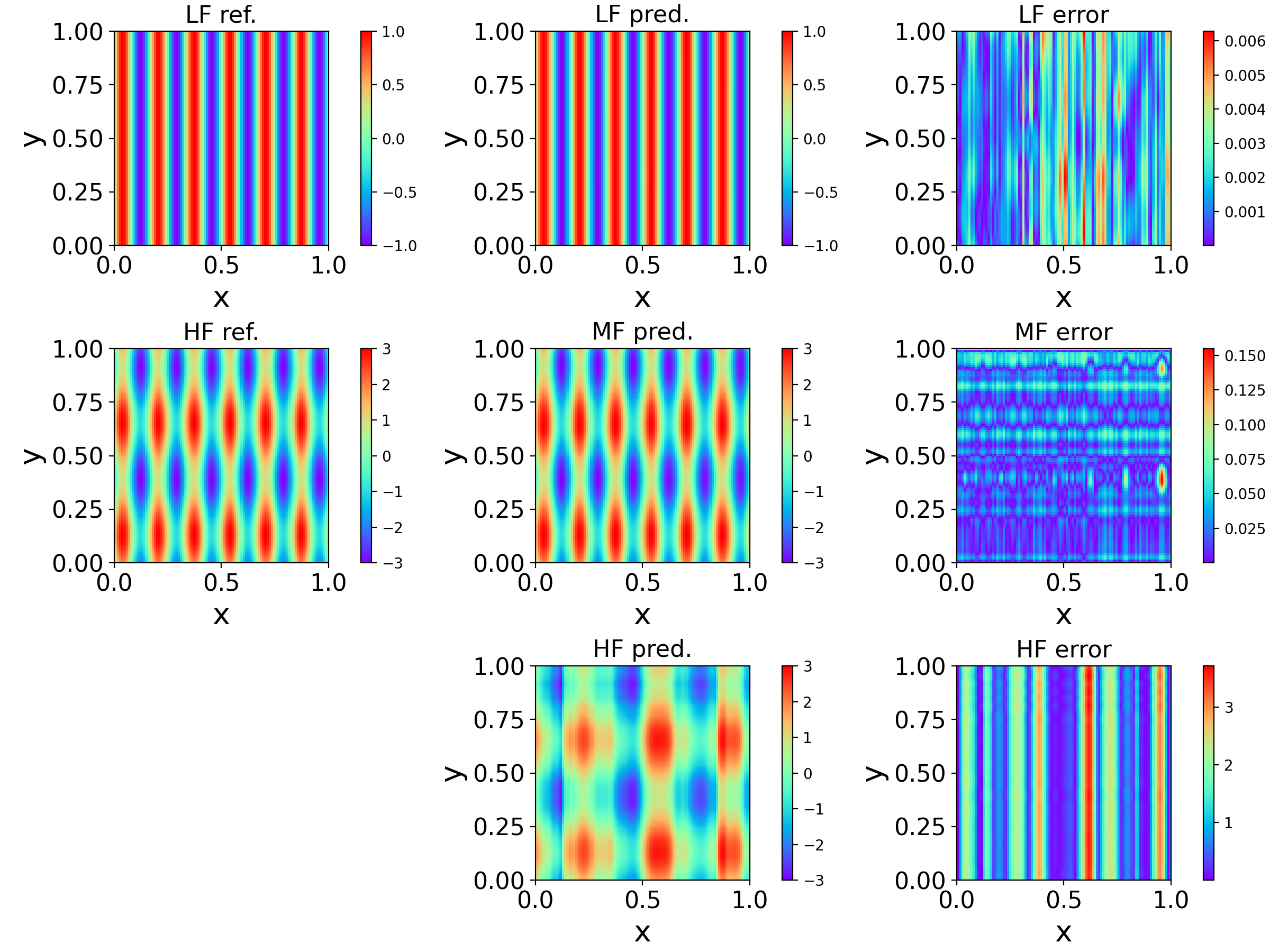}
    \caption{Results for Test 3. Top row: low-fidelity reference solution and prediction. Middle row: multifidelity reference solution and prediction. Bottom row: high-fidelity prediction.}
    \label{fig:Test3b}
\end{figure}

Plots of the high-fidelity and multifidelity predictions of the high-fidelity data are given in Fig. \ref{fig:Test3a}c and d. Without the low-fidelity data, the high-fidelity prediction is unable to capture the high-frequency oscillations in the $x$-direction. In contrast, the multifidelity prediction agrees well with the reference solution. The full plots of the solutions are given in Fig. \ref{fig:Test3b}. The multifidelity prediction has a much smaller absolute error than the high-fidelity prediction (Fig. \ref{fig:Test3b}, right column.) The relative $\ell_2$ error of the low-fidelity prediction is 0.0033, and the relative $\ell_2$ errors of the multifidelity and high-fidelity predictions are 0.0201 and 0.9641, respectively. 


\subsection{Test 4: Higher-dimensional problem}\label{sec:Test4}
\begin{figure}[h]
    \centering
    \includegraphics[width=\textwidth]{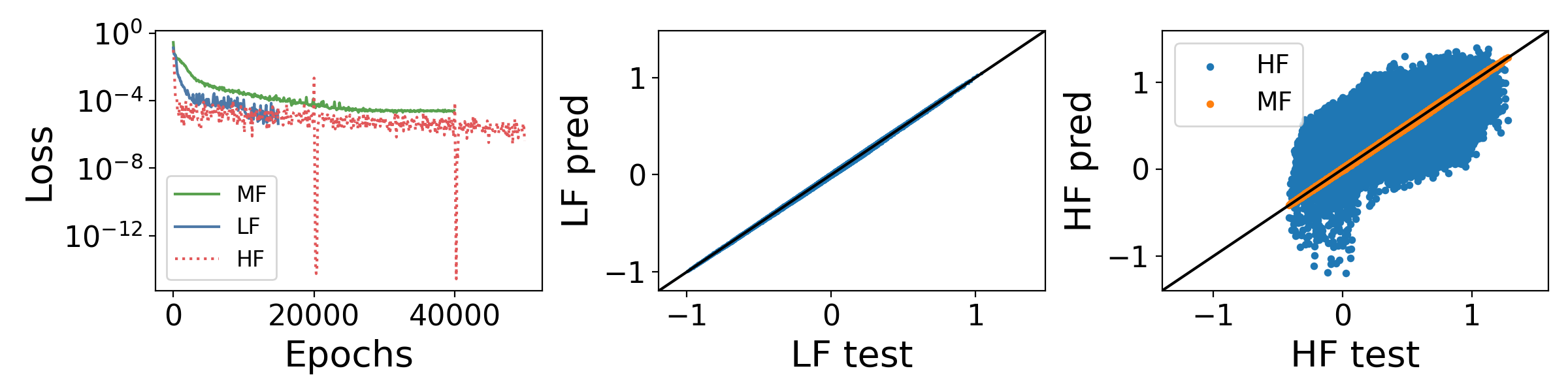}\\
    \includegraphics[width=\textwidth]{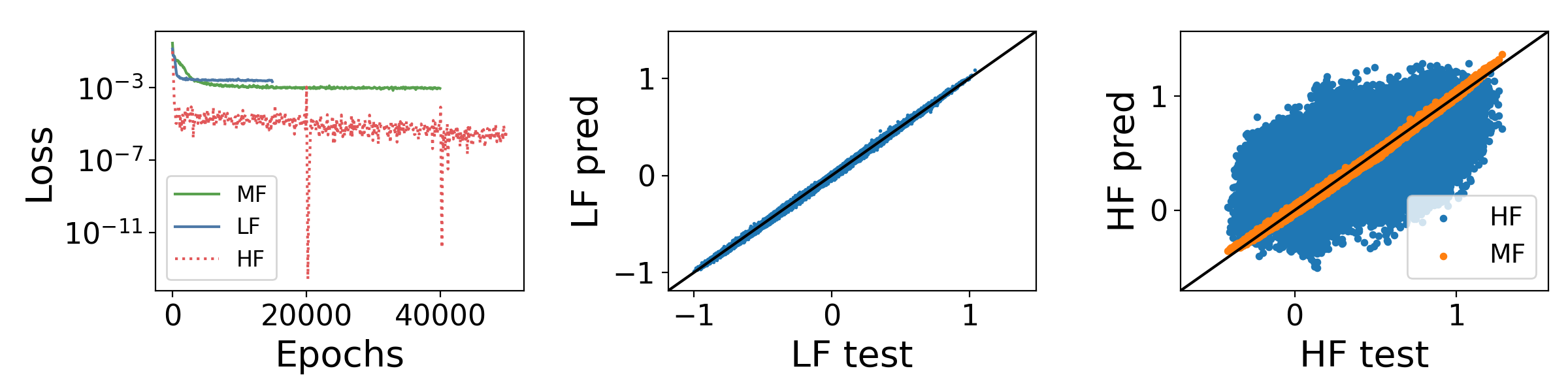}
    \caption{Results for Test 4 with clean data (top) and noisy data (bottom). Noisy data has white noise with variance $\sigma_L = 0.05$ added to the low-fidelity data and $\sigma_H = 0.03$ added to the high-fidelity data. Left to right: loss curves, low-fidelity prediction, and high-fidelity prediction.}
    \label{fig:Test4}
\end{figure}

We now turn to a higher-dimensional problem:
\begin{align}
    f_L(x_1, x_2, x_3, x_4) &= 1.2f_H(x_1, x_2, x_3, x_4)-0.5, \\
    f_H(x_1, x_2, x_3, x_4) &=  0.5(0.1 \exp(x_1+x_2)-x_4 \sin(12 \pi x_3) + x_3),
\end{align}
for $x_i \in [0, 1]$, $i=1, 2, 3, 4.$
To further test the method, we also consider a case with Gaussian white noise added to both the low-fidelity and high-fidelity training data. The variance of the noise is denoted by $\sigma_L$ and $\sigma_H$ for the low- and high-fidelity samples, respectively. To generate the training data we take $N_{LF} = 25,000$ and $N_{HF}=150$ and sample the training points randomly in $\Omega = [0, 1]^4$. We show the results of training in Table \ref{tab:error-test4} and Fig. \ref{fig:Test4}. The high-fidelity predictions have large relative errors and are not able to capture the true solution. In contrast, the multifidelity predictions are much more accurate, even in the presence of noisy data. The errors do increase somewhat when noise is added to the training data.

\begin{table}[h]
    \centering
    \begin{tabular}{l | c  c  c } 
     \hline \hline
      & LF & MF & HF \\ \hline
$\sigma_L = 0$, $\sigma_H = 0$ &  0.0207 & 0.0110 & 0.5046 \\
$\sigma_L = 0.05$, $\sigma_H = 0.03$ & 0.0408 & 0.0301 & 0.5297 \\
     \hline     \hline
    \end{tabular}
        \caption{Relative $\ell_2$ errors for Test 4.}
    \label{tab:error-test4}
\end{table}

\subsection{Test 5: Physics-informed training}\label{sec:Test5}
Physics-informed KANs (PIKANs) have attracted recent attention \cite{rigas2024adaptive, ranasinghe2024ginn, shukla2024comprehensive, wang2024kolmogorov, patra2024physics, shuai2024physics, sohigher}. PIKANs use the basic ideas of physics-informed neural networks (PINNs) \cite{raissi2019physics}, but with KANs instead of MLPs as the neural architecture. One can consider many cases in which physics-informed training could be useful within a multifidelity framework. Perhaps the available data is on a coarse mesh but the physics of the system is known. In this case, physics-informed training can be used instead of high-fidelity data to perform something akin to super-resolution (\textit{e.g.,} \cite{gao2021super, zayats2022super, yang2024super, cai2024physics, aliakbari2022predicting}). In other cases, a small amount of high-fidelity data may be available, as well as low-order physics-based models. In these cases, we can use physics instead of the low-fidelity data, which is then corrected with the high-fidelity data. A third use case arises from the observations in \cite{shukla2024comprehensive, rigas2024adaptive, patra2024physics} that PIKANs can be difficult to train. In practice, single-fidelity PIKANs can fail to accurately capture the true solution of the equation used to train the KAN. In these cases, we can use physics at both levels of the multifidelity training, to correct a solution that is learned with too high an error, along the lines of iteratively learning a solution that has recently become popular for PINNs \cite{howard2023stacked, wang2024multi, aldirany2024multi, ainsworth2021galerkin, ainsworth2022galerkin}. It is an example of this form that we consider in this section. 

\begin{figure}[h]
    \centering
    \includegraphics[width=0.7\textwidth]{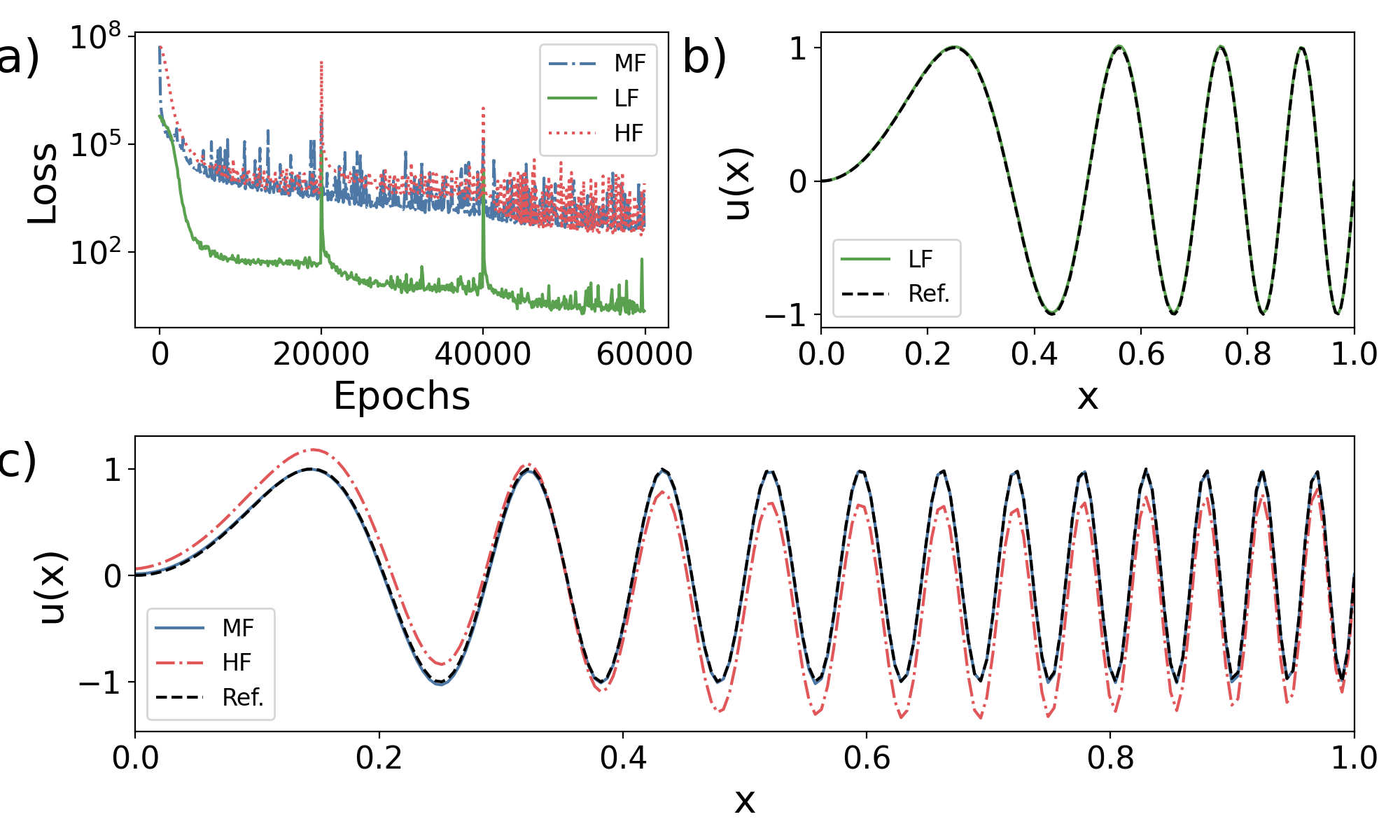}
    \caption{Results for Test 5. a) Loss curves. b) Low-fidelity reference solution and prediction. c) High-fidelity reference solution and predictions.}
    \label{fig:Test5}
\end{figure}

We consider the Poisson equation
\begin{align}
    \frac{d^2u}{dx^2} &= f(x), \; x \in [0, 1], \\
    u(0)&=u(1) = 0,
\end{align}
with exact solution $u(x) = \sin(2k\pi x^2),$ which determines $f(x)$. For the target high-fidelity equation we use $k=12$, and for the low-fidelity we use $k=4$. In this case we train without data, and instead use a physics-informed loss function \cite{raissi2019physics}. Results are shown in Fig. \ref{fig:Test5}. The high-fidelity KAN attempts to predict the solution to the Poisson equation with $k=12$. However, it has a large relative $\ell_2$ error of 0.346. In contrast, the multifidelity prediction predicts the low-fidelity reference solution with a relative error of 0.0153, and the high-fidelity reference solution with a relative error of 0.0221.

\subsection{Test 6: Mechanical MNIST}\label{sec:Test6}

We consider the multifidelity Mechanical MNIST dataset \cite{MMnistMF, lejeune2020mechanical}. The dataset is created by applying uniaxial extension to materials created from MNIST \cite{deng2012mnist}. We consider two meshes as our low- and high-fidelity data. The high-fidelity dataset has a fully refined mesh with quadratic triangular elements (denoted by UE in \cite{lejeune2021exploring}). The low-fidelity dataset is generated using  $14 \times 14 \times 2$ linear triangular elements (UE-CM-14). We train with the 60,000 samples from the low-fidelity training set, and $N$ samples from the high-fidelity dataset. Then, we test with 10,000 samples from the low- and high-fidelity datasets. 

The goal is to predict the total change in strain energy when the block is stretched to 50\% of its original length, denoted by $\nabla \psi$. From \cite{lejeune2020mechanical}, the high-fidelity mechanical MNIST dataset trained with an MLP results a mean percent error (MPE) of 2.5\% when trained with the entire high-fidelity dataset \cite{lejeune2020mechanical} and the MPE is reduced to 1.0\% with a convolutional neural network (CNN) \cite{lejeune2021exploring} (we emphasize that we use the MPE instead of the relative $\ell_2$ error in this section to align with \cite{lejeune2020mechanical}).  Using transfer learning with a low-fidelity dataset can further reduce the MSE and reduce the amount of high-fidelity data needed to train \cite{lejeune2021exploring}.

We first test the training of the KAN on the mechanical MNIST high-fidelity dataset to compare the MPE between MLPs and KANs. We note that while CNNs have shown better performance for MNIST, a multifidelity convolutional KAN extending \cite{bodner2024convolutional} is outside the scope of this work. As a baseline, the KAN trained with the entire high-fidelity dataset achieves an MPE of 1.26\%. 

\begin{figure}[h]
    \centering
    \includegraphics[width=0.9\textwidth]{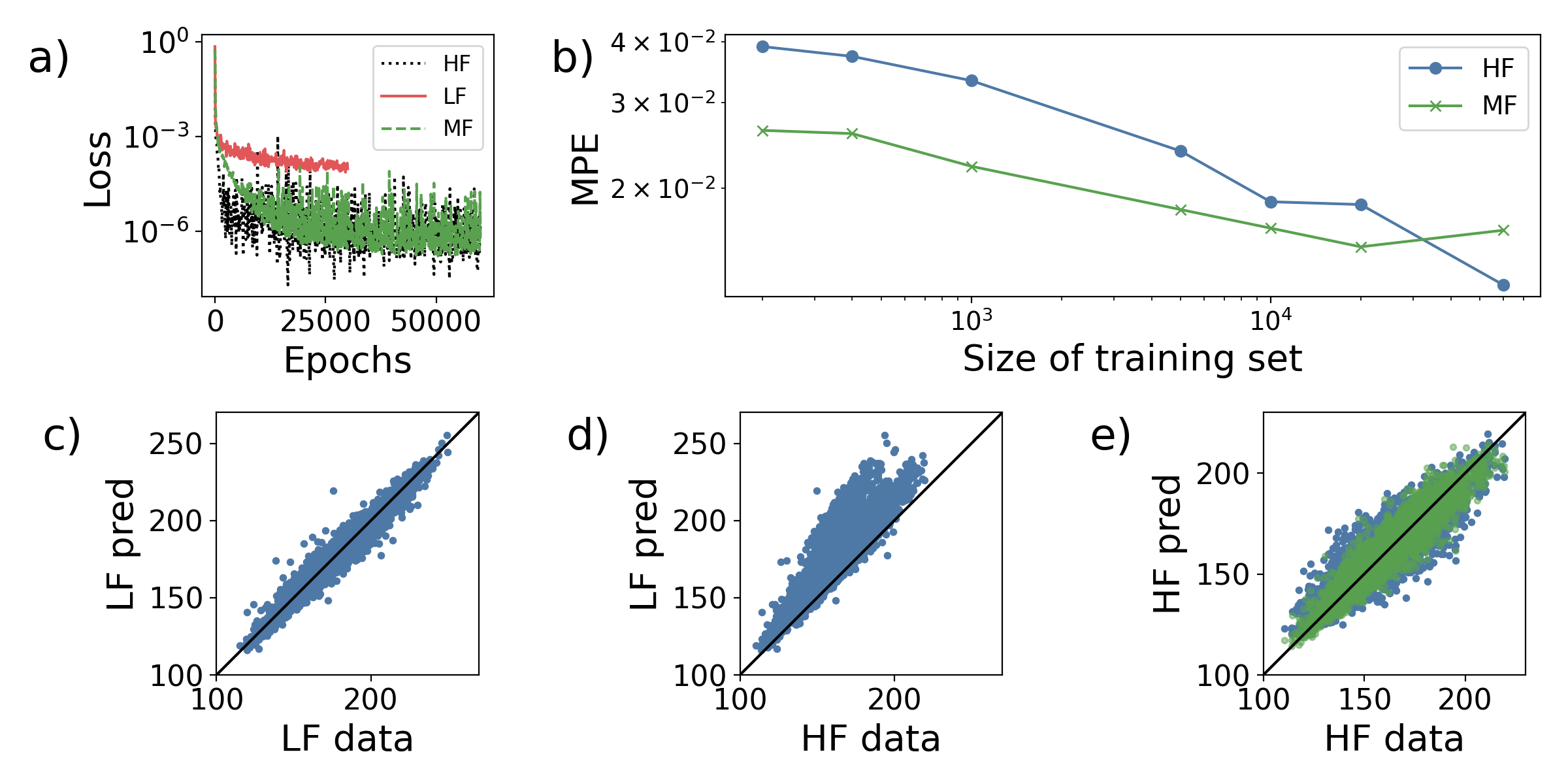}
    \caption{Results for Test 6.  a) Loss curves. b) MPE for both MFKAN and high-fidelity KAN tests. c) LF predictions of the LF data compared with reference LF data. d) LF predictions of the HF data, compared with reference HF data. e) HF predictions of the HF data, compared with reference HF data for the high-fidelity KAN (blue) and the MFKAN (green). }
    \label{fig:Test6}
\end{figure}

The low-fidelity model has an MPE of 1.60\% on the low-fidelity test set and 8.53\% on the high-fidelity test set. Clearly, the low-fidelity KAN is a good model for the low-fidelity test set, but it does not accurately predict the high-fidelity data. We test the high-fidelity KAN and MFKAN by varying the size of the high-fidelity training set from 200 to 60,000. As shown in Fig. \ref{fig:Test6}b, the MFKAN has a lower test MPE than the high-fidelity KAN in all cases except when trained with the full training set with 60,000 samples. This demonstrates that in the presence of limited high-fidelity data, the MFKAN can outperform a KAN trained with only high-fidelity data. 

\subsection{Test 7: Multifidelity extrapolation}\label{sec:Test7}

To test our multifidelity method with a more realistic application, we consider the case of developing a surrogate for an expensive numerical code. Neural architectures, in general, struggle with extrapolation, or predictions outside the domain of input values. This means that for models trained with numerical data, we must have numerical data spanning the entire time range at which we want predictions, which can be very expensive to produce. In this section, we demonstrate that we can successfully extrapolate beyond our high-fidelity data if we have a low-fidelity numerical code for generating low-fidelity data. 

Specifically, we consider the vorticity field of a fluid flow past a cylinder generated with Phi Flow \cite{holl2024phiflow} to match the Wake Flow tutorial. The dimensionless domain is $\Omega = [0, 200]\times[0, 100]$, with a cylinder of radius 10 located at $(20, 50)$. For the high-fidelity data, we use a $257 \times 129$ mesh. We select 200 consecutive time steps after the unsteady flow has developed. For the low-fidelity data, we sub-sample the high-fidelity mesh to create a $86 \times 43$ mesh, keeping the same time steps. Importantly, we train with all the low-fidelity data and only the first 100 timesteps of the high-fidelity data. This means that when predicting for $t>100$, the model is extrapolating in time with respect to the high-fidelity data.

Instead of training a single-fidelity KAN for the low-fidelity data, we directly use our numerical solver to generate the low-fidelity input to the multifidelity network. This reduces the error that can occur due to the low-fidelity prediction.

\begin{figure}[h]
    \centering
    \includegraphics[width=0.9\textwidth]{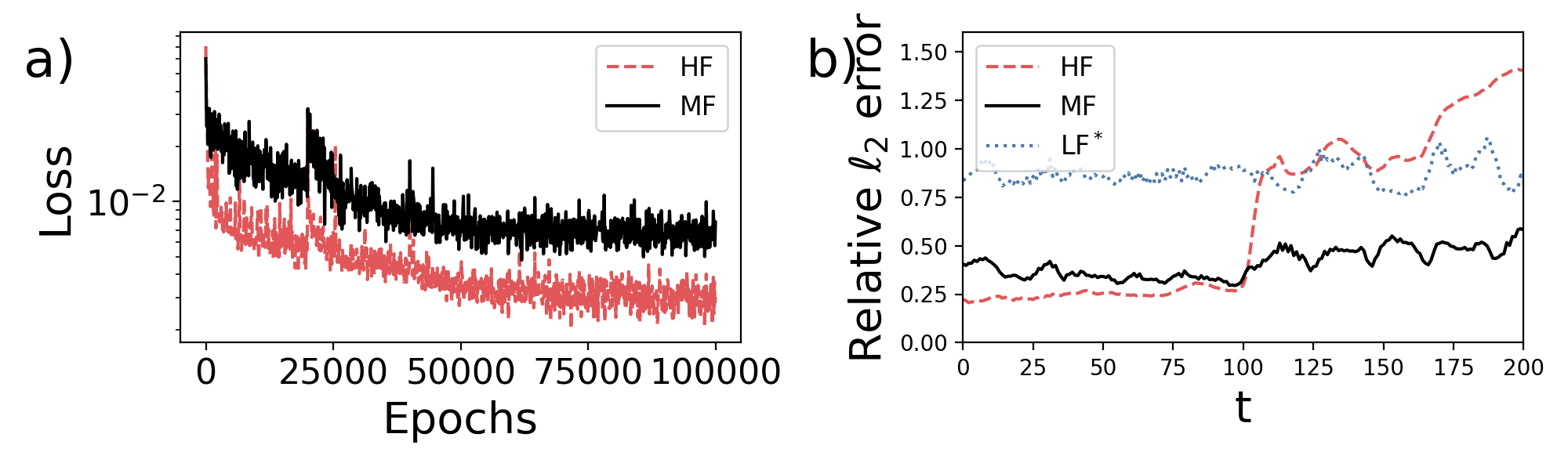}
    \caption{Results for Test 7. a) Loss curves. b) Relative $\ell_2$ error over time for the high-fidelity and multifidelity predictions. The low-fidelity error is the error of the low-fidelity data extrapolated to the high-fidelity mesh.}
    \label{fig:Test7a}
\end{figure}

In Figure \ref{fig:Test7a}b we show the relative $\ell_2$ error over time for the multifidelity and high-fidelity predictions. In the area where there is no high-fidelity data, $t\in[100,200]$, the high-fidelity relative $\ell_2$ error is significantly higher, after jumping rapidly at $t=100$. In contrast, the multifidelity error increases only slightly in the extrapolation region and does not grow substantially with time. The multifidelity model has a significantly lower error than just interpolating the low-fidelity data to the high-fidelity mesh. 

\begin{figure}[h]
    \centering
    \includegraphics[width=0.9\textwidth]{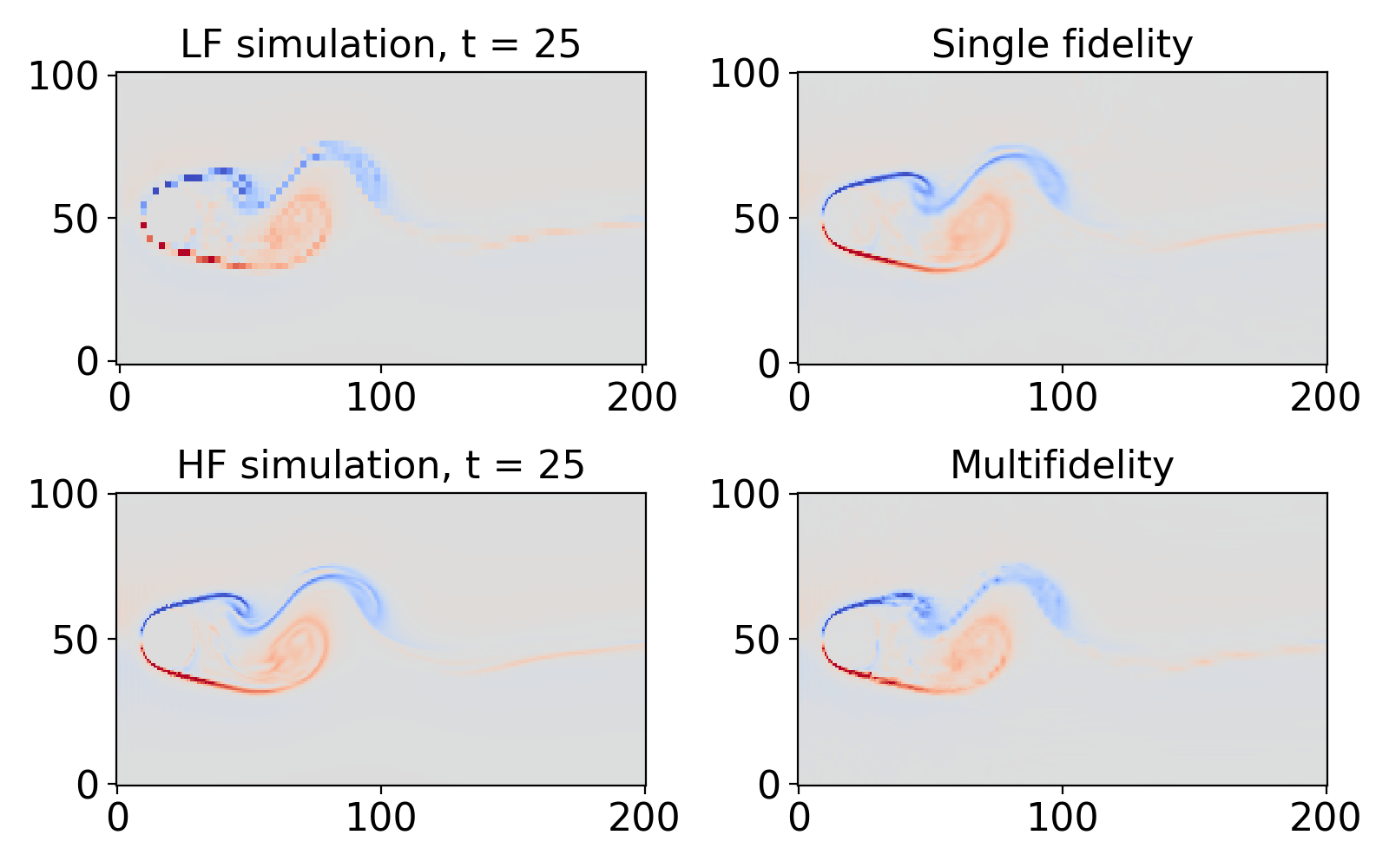}\\
    \includegraphics[width=0.9\textwidth]{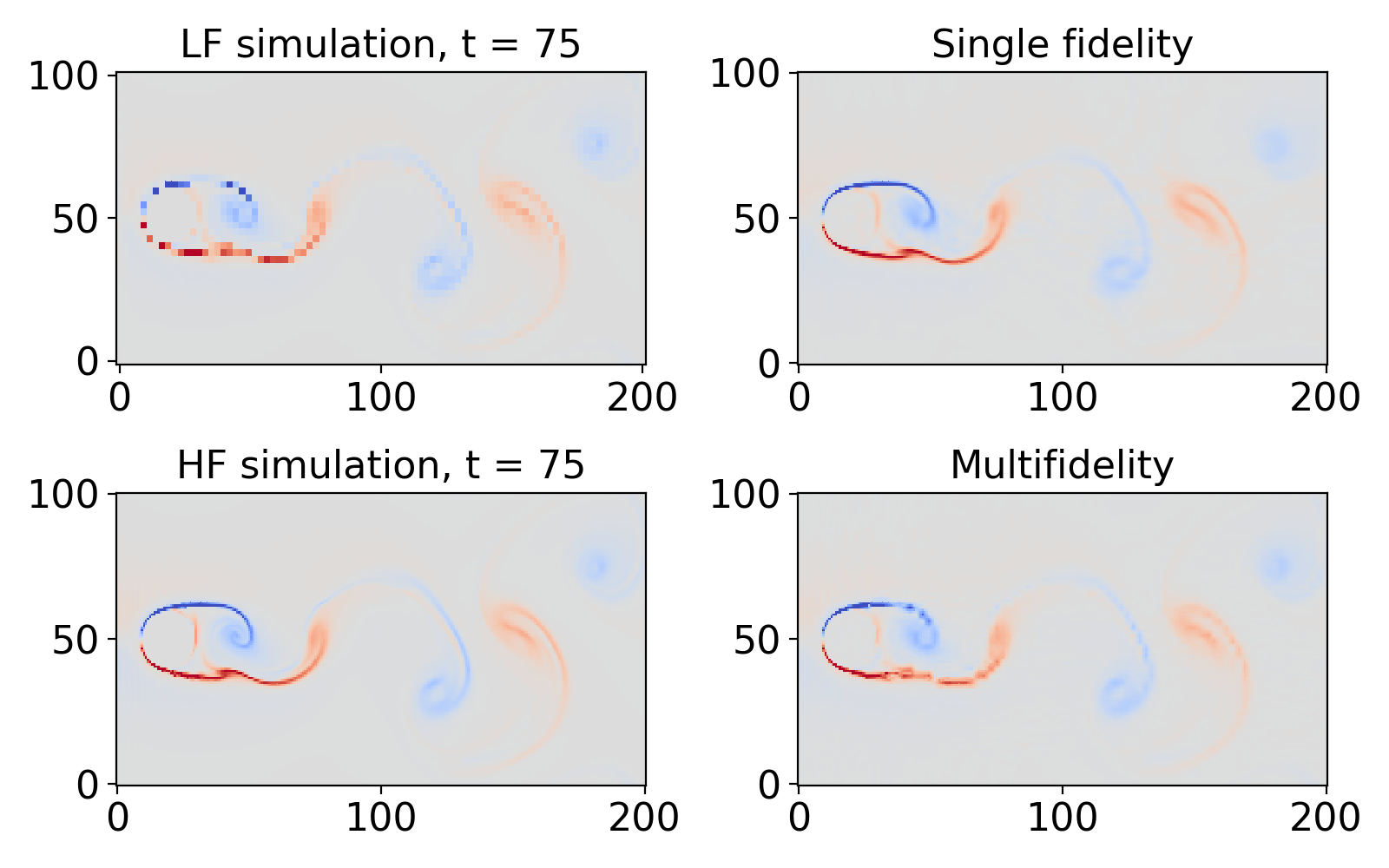}
    \caption{Results for Test 7 in the time-extrapolation domain at $t=25$ and $t=75$. Top left: low-fidelity simulation output. Bottom left: high-fidelity simulation output. Top right: single-fidelity prediction. Bottom right: multifidelity prediction.}
    \label{fig:Test7b}
\end{figure}

\begin{figure}[h]
    \centering
    \includegraphics[width=0.9\textwidth]{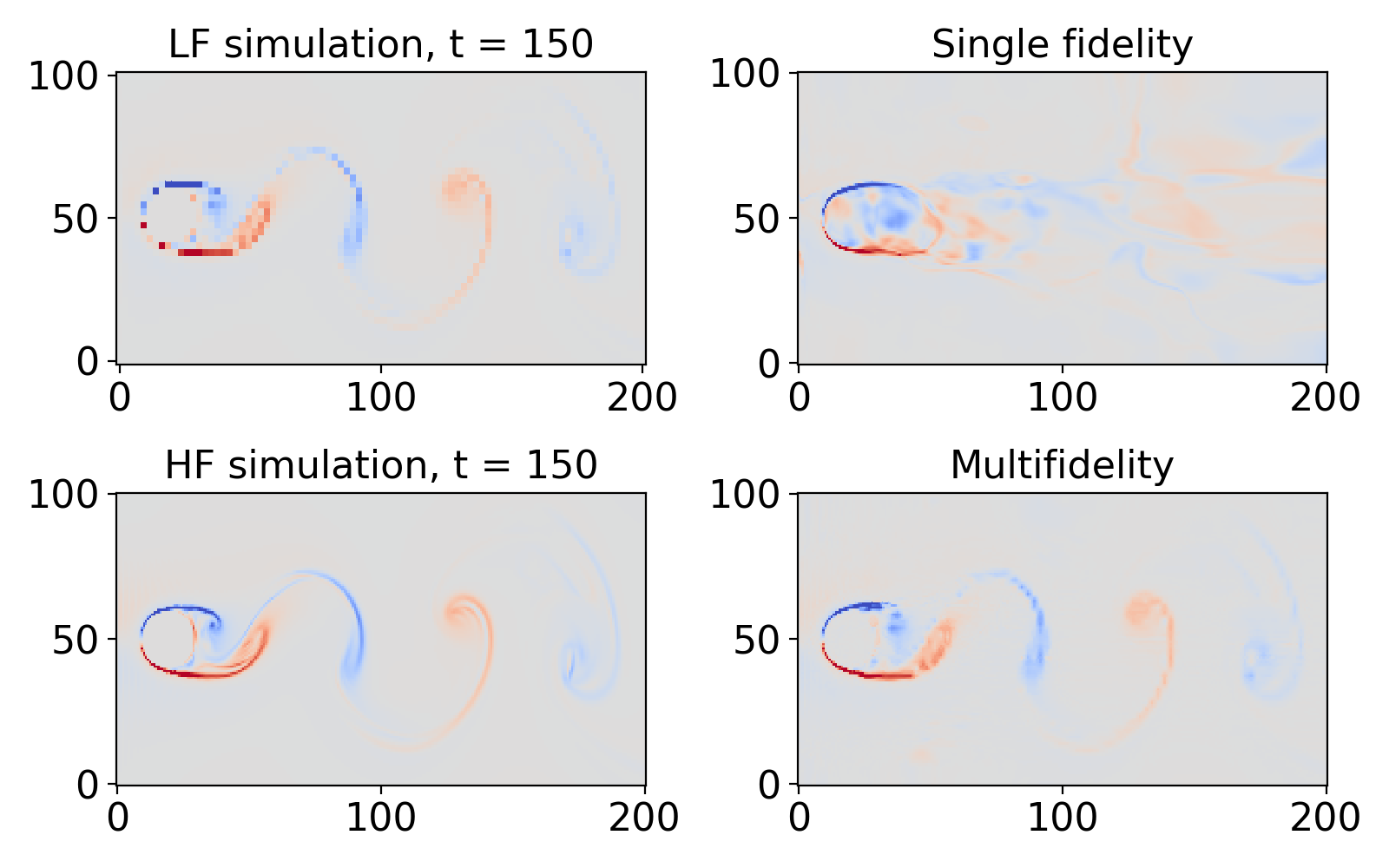}\\
    \includegraphics[width=0.9\textwidth]{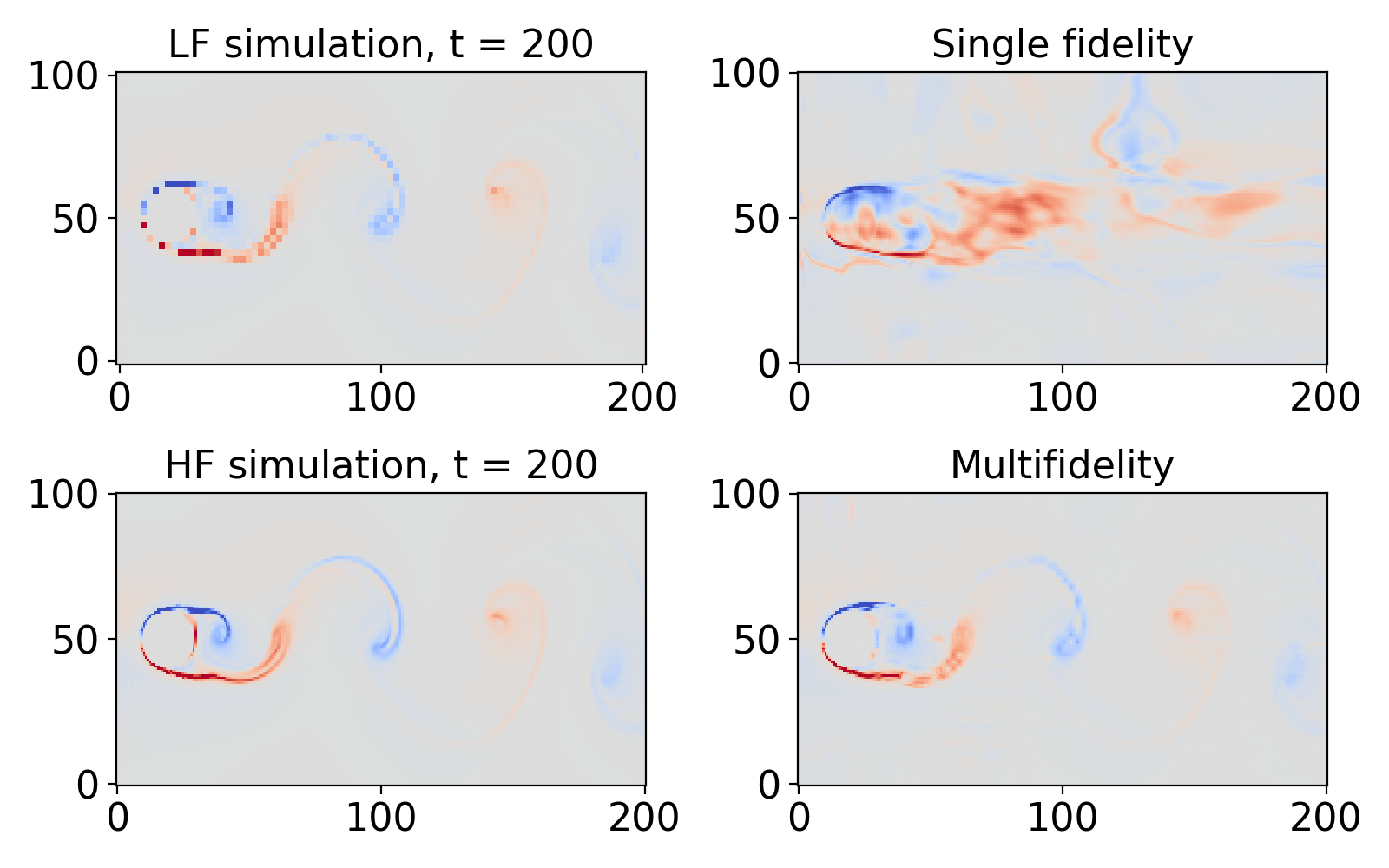}
    \caption{Results for Test 7 in the time-extrapolation domain at $t=150$ and $t=200$. Top left: low-fidelity simulation output. Bottom left: high-fidelity simulation output. Top right: single-fidelity prediction. Bottom right: multifidelity prediction.}
    \label{fig:Test7c}
\end{figure}

In Figs. \ref{fig:Test7b} and \ref{fig:Test7c} we show sample outputs from the simulations and model predictions. Fig. \ref{fig:Test7b} contains two dimensionless time snapshots for the time domain where high-fidelity data is provided. Both models are able to capture the high-fidelity simulation data, although the high-fidelity model does have slightly higher accuracy. Fig. \ref{fig:Test7c} has two snapshots in the extrapolation region. The single-fidelity solution no longer captures the vortex structure accurately. In contrast, the multifidelity model remains accurate at $t=200.$

This example shows the potential power of multifidelity modeling for reducing the computational cost of generating data for machine-learned surrogates. While a small amount of high-fidelity data is still necessary, the model can extrapolate forward in time given only additional low-fidelity data, which is less expensive to produce.

%\subsection{Application: learning from noisy data}

%To further demonstrate the capability of our proposed method, we solve a steady, laminar flow past a circular cylinder using the analytical solution of a steady-state Poiseuille flow as the low-fidelity data.

\section{Conclusions}
We have developed an adaptable, robust framework for multifidelity modeling with KANs. Our method takes advantage of the B-splines learned in KANs to learn the nonlinear and linear correlations between the low- and high-fidelity datasets. In particular, this method does not require a nested dataset (that is, the high-fidelity dataset does not need to be a subset of the low-fidelity dataset) because a low-fidelity KAN is trained on the low-fidelity dataset, which then serves as a surrogate for the low-fidelity data. This flexibility allows the method to be adapted to a variety of complex applications, where nested datasets may not be available.

As noted in the introduction, the field of training KANs has been rapidly changing, with many variations being added weekly and monthly. We note that the multifidelity framework presented here does not rely on KANs as implemented in \cite{liu2024kan}. In particular, the nonlinear and low-fidelity blocks can be replaced by KAN variants as they emerge, including recent advances in different basis functions \cite{ss2024chebyshev, seydi2024exploring, yu2024sinc, reinhardt2024sinekan, ta2024fc}. Additionally, the flexibility of the method allows for different network types to be used for each block to best fit the data. For instance, an MLP could be used for the low-fidelity data and a KAN for the nonlinear network. This flexibility presents opportunities for adapting the framework for the best performance as KANs continue to develop and for modifying the method to best fit the characteristics of a given dataset.

\section{Data and code availability}
The Mechanical MNIST dataset in Sec. \ref{sec:Test6} is available at \cite{MMnistMF}. The fluid flow data in Sec. \ref{sec:Test7} was produced with \cite{holl2024phiflow}. The KANs results in this work were produced with code adapted from JaxKAN \cite{Rigas_jaxKAN_A_JAX-based_2024, rigas2024adaptive}. All code and data will be released upon publication in a journal. For those wishing to implement the method themselves, tutorials for some examples are available in Neuromancer \cite{Neuromancer2023} at \href{https://github.com/pnnl/neuromancer/blob/feature/mfkans/examples/KANs/p3_mfkan_example_1d.ipynb}{https://github.com/pnnl/neuromancer/blob/feature/mfkans/examples/KANs/}. Training parameters used to train the results in this paper are given in Sec. \ref{app:training}.

\section{Acknowledgements}
The code to plot the KAN diagrams in the graphical abstract and \cref{fig:MF_KAN} was adapted from \texttt{pykan} \cite{liu2024kan}. 

This project was completed with support from the U.S. Department of Energy, Advanced Scientific Computing Research program, under the Scalable, Efficient and Accelerated Causal Reasoning Operators, Graphs and Spikes for Earth and Embedded Systems (SEA-CROGS) project (Project No. 80278) and under the Uncertainty
Quantification for Multifidelity Operator Learning (MOLUcQ) project (Project No. 81739). The computational work was performed using PNNL Institutional Computing at Pacific Northwest National Laboratory. Pacific Northwest National Laboratory (PNNL) is a multi-program national laboratory operated for the U.S. Department of Energy (DOE) by Battelle Memorial Institute under Contract No. DE-AC05-76RL01830.

\clearpage

\bibliographystyle{unsrt}
\bibliography{references}  %%% Uncomment this line and comment out the ``thebibliography'' section below to use the external .bib file (using bibtex) .

\appendix
\section{Training parameters} \label{app:training}
All results in this paper are implemented in \texttt{JAX} \cite{jax2018github} using the \texttt{Jax-KAN} \cite{Rigas_jaxKAN_A_JAX-based_2024} KAN implementation. All networks are trained with the \texttt{ADAM} optimizer. 

Boundaries are the iterations at which the grid is refined, following the procedure in \cite{Rigas_jaxKAN_A_JAX-based_2024}. When the grid is refined, the learning rate is scaled using the given scales. If no scales are given, the grid is kept fixed. 

\subsection{Test 1}
% Suggested design
\begin{table}[h]
    \centering
    \begin{tabular}{c  c } 
     \hline\hline
     Parameter &  \\ 
     \hline\hline
    Low-fidelity KAN architecture &  [1, 5, 1]   \\ 
    $g_{L}$   &       [5, 10, 15]                     \\ 
    Low-fidelity scales & [1, .6, .6] \\
    Low-fidelity boundaries & [0, 5000, 1000] \\
    $k_{L}$   &          3              \\ 
    Low-fidelity learning rate &      0.001        \\ 
    Low-fidelity iterations    &  15001             \\ 
    $N_{LF}$    &          50 and 300     \\ 
    \hline
    Nonlinear KAN architecture &   [2, 4, 1]   \\ 
    $g_{nl}$   &         3                   \\ 
    $k_{nl}$   &          2              \\ 
    Linear KAN architecture &    [2, 1]   \\ 
    $g_{l}$   &            1                \\ 
    $k_{l}$   &            1             \\ 
    
    High-fidelity learning rate &       0.005      \\ 
    High-fidelity iterations    &    10001        \\ 
    $N_{HF}$    &       5        \\ 
    $w$ &  0 and 10 \\
    $n$ & 2\\
    $\lambda_{\alpha}$ & 10 \\
     \hline

    Single-fidelity KAN architecture &    [1, 2, 1] \\ 
    $g$   &     2                      \\ 
    $k$   &      3                  \\ 
    Single-fidelity learning rate &       0.005       \\ 
    Single-fidelity iterations    &       10001        \\ 
    \hline
    
    \end{tabular}
    \caption{Hyperparameters used for the results in~\cref{sec:Test1}.}
    \label{tab:params_Test1}
\end{table}
\clearpage

\subsection{Test 2}
% Suggested design
\begin{table}[h]
    \centering
    \begin{tabular}{c  c } 
     \hline\hline
     Parameter &  \\ 
     \hline\hline
    Low-fidelity KAN architecture &  [1, 5,  1]   \\ 
    $g_{L}$   &       [6, 12]                    \\ 

    $k_{L}$   &          3              \\ 
    Low-fidelity learning rate &      0.001        \\ 
    Low-fidelity iterations    &      40001         \\ 
        Low-fidelity scales & [1, 0.4] \\
    Low-fidelity boundaries & [0, 15000] \\
    $N_{LF}$    &       51        \\ 
    \hline
    Nonlinear KAN architecture &  [2, 8, 1]    \\ 
    $g_{nl}$   &     [5, 8]                       \\ 
    $k_{nl}$   &        2                \\ 

    Linear KAN architecture &   [2, 1]   \\ 
    $g_{l}$   &            1                \\ 
    $k_{l}$   &            1             \\ 
    
    High-fidelity learning rate &       0.005       \\ 
    High-fidelity iterations    &    100001        \\ 
    High-fidelity scales & [1, 0.7] \\
    High-fidelity boundaries & [0, 25000] \\
    $N_{HF}$    &      14         \\ 
    $w$ & 0 and 1 \\
    $n$ & 4 \\
    $\lambda_{\alpha}$ & 0.01 \\
     \hline

         Single-fidelity KAN architecture &    [1, 5, 1] \\ 
    $g$   &     6                      \\ 
    $k$   &      3                  \\ 
    Single-fidelity learning rate &       0.001       \\ 
    Single-fidelity iterations    &       50001        \\ 
    \hline
    
    \end{tabular}
    \caption{Hyperparameters used for the results in~\cref{sec:Test2}.}
    \label{tab:params_Test2}
\end{table}
\clearpage

\subsection{Test 3}
% Suggested design
\begin{table}[h]
    \centering
    \begin{tabular}{c  c } 
     \hline\hline
     Parameter &  \\ 
     \hline\hline
    Low-fidelity KAN architecture &  [2, 10, 1]   \\ 
    $g_{L}$   &               6            \\ 
    $k_{L}$   &               3         \\ 
    Low-fidelity learning rate &      0.001        \\ 
    Low-fidelity iterations    &       30000        \\ 
    $N_{LF}$    &        10000       \\ 
    \hline
    Nonlinear KAN architecture &  [3, 10, 1]    \\ 
    $g_{nl}$   &                   5         \\ 
    $k_{nl}$   &                 3       \\ 
    Linear KAN architecture & [3, 1]     \\ 
    $g_{l}$   &            1                \\ 
    $k_{l}$   &            1             \\ 
    
    High-fidelity learning rate &     0.008         \\ 
    High-fidelity iterations    &     50000       \\ 
    $N_{HF}$    &          150     \\ 
    $w$ &  0.001 \\
    $n$ & 4 \\
    $\lambda_{\alpha}$ & 10 \\
     \hline
     
         Single-fidelity KAN architecture &    [2, 10, 1] \\ 
    $g$   &     5                      \\ 
    $k$   &      3                  \\ 
    Single-fidelity learning rate &       0.008       \\ 
    Single-fidelity iterations    &       50000        \\ 
    \hline

    \end{tabular}
    \caption{Hyperparameters used for the results in~\cref{sec:Test3}.}
    \label{tab:params_Test3}
\end{table}
\clearpage

\subsection{Test 4}
% Suggested design
\begin{table}[h]
    \centering
    \begin{tabular}{c  c } 
     \hline\hline
     Parameter &  \\ 
     \hline\hline
    Low-fidelity KAN architecture &   [4, 10, 1]  \\ 
    $g_{L}$   &          [6, 12]                 \\ 
    $k_{L}$   &      3                  \\ 

    Low-fidelity learning rate &  0.005            \\ 
    Low-fidelity iterations    &  15000             \\ 
        Low-fidelity scales & [1, .7] \\
    Low-fidelity boundaries & [0,  10000] \\
    $N_{LF}$    &     25000          \\ 
    \hline
    Nonlinear KAN architecture &  [5, 6,  1]    \\ 
    $g_{nl}$   &          5                  \\ 
    $k_{nl}$   &           3             \\ 
    Linear KAN architecture &    [5,   1]  \\ 
    $g_{l}$   &            1                \\ 
    $k_{l}$   &            1             \\ 
    
    High-fidelity learning rate &    0.008          \\ 
    High-fidelity iterations    &      40000      \\ 
    $N_{HF}$    &      150         \\ 
    $w$ &  1 \\
    $n$ & 4 \\
    $\lambda_{\alpha}$ & 10 \\
     \hline

              Single-fidelity KAN architecture &   [4, 10,  1] \\ 
    $g$   &               [5, 8, 10]            \\ 
    $k$   &      3                  \\ 

    Single-fidelity learning rate &     0.008      \\ 
    Single-fidelity iterations    &    50000     \\ 
        Single-fidelity scales & [1, .8, .8] \\
    Single-fidelity boundaries & [0,  20000, 40000] \\
    \hline

    \end{tabular}
    \caption{Hyperparameters used for the results in~\cref{sec:Test4}.}
    \label{tab:params_Test4}
\end{table}
\clearpage

\subsection{Test 5}
% Suggested design
\begin{table}[h]
    \centering
    \begin{tabular}{c  c } 
     \hline\hline
     Parameter &  \\ 
     \hline\hline
    Low-fidelity KAN architecture &   [1, 10, 1]  \\ 
    $g_{L}$   &            [6, 12, 18]               \\ 
    $k_{L}$   &         3               \\ 

    Low-fidelity learning rate &      0.0005        \\ 
    Low-fidelity iterations    &       60000        \\ 
            Low-fidelity scales & [1, .8, .8] \\
    Low-fidelity boundaries & [0, 20000, 40000] \\
    $N_{LF}$    &     1000          \\ 
    \hline
    Nonlinear KAN architecture &  [2, 10, 10, 1]    \\ 
    $g_{nl}$   &       [6, 12, 18]                     \\ 
    $k_{nl}$   &       3                 \\ 
    Linear KAN architecture &   [2, 1]   \\ 
    $g_{l}$   &            1                \\ 
    $k_{l}$   &            1             \\ 
    
    High-fidelity learning rate &    0.01          \\ 
    High-fidelity iterations    &   60000         \\ 
        High-fidelity scales & [1, .8, .8] \\
    High-fidelity boundaries & [0, 20000, 40000] \\
    $N_{HF}$    &      1000         \\ 
    $w$ & 0  \\
    $n$ & 4 \\
    $\lambda_{\alpha}$ & 100000 \\
     \hline

                   Single-fidelity KAN architecture &   [1, 8, 8, 1] \\ 
    $g$   &      [6, 12, 18]                     \\ 
    $k$   &      3                  \\ 
    Single-fidelity learning rate &     0.0005      \\ 
    Single-fidelity iterations    &       60000  \\ 
    Single-fidelity scales & [1, .8, .8] \\
    Single-fidelity boundaries & [0, 20000, 40000] \\
    \hline
    \end{tabular}
    \caption{Hyperparameters used for the results in~\cref{sec:Test5}.}
    \label{tab:params_Test5}
\end{table}
\clearpage

\subsection{Test 6}
% Suggested design
\begin{table}[h]
    \centering
    \begin{tabular}{c  c } 
     \hline\hline
     Parameter &  \\ 
     \hline\hline
    Low-fidelity KAN architecture &   [784, 64, 1]  \\ 
    $g_{L}$   &                 5          \\ 
    $k_{L}$   &     3                   \\ 
    Low-fidelity learning rate &        0.01      \\ 
    Low-fidelity iterations    &    30001           \\ 
    $N_{LF}$    &    60000           \\ 
    \hline
    Nonlinear KAN architecture &   [785, 32, 1]   \\ 
    $g_{nl}$   &       5                     \\ 
    $k_{nl}$   &        3                \\ 
    Linear KAN architecture &    [1, 1]  \\ 
    $g_{l}$   &            1                \\ 
    $k_{l}$   &            1             \\ 
    
    High-fidelity learning rate & 0.01             \\ 
    High-fidelity iterations    &   60001         \\ 
    $w$ & 0\\
    $n$ & 4\\
    $\lambda_{\alpha}$ & 0.01\\
     \hline

     Single-fidelity KAN architecture &  [784, 64, 1]  \\ 
    $g$   &         5                  \\ 
    $k$   &      3                  \\ 
    Single-fidelity learning rate &        0.01   \\ 
    Single-fidelity iterations    &     60001    \\ 
         \hline

    \end{tabular}
    \caption{Hyperparameters used for the results in~\cref{sec:Test6}.}
    \label{tab:params_Test6}
\end{table}
\clearpage

\subsection{Test 7}
% Suggested design
\begin{table}[h]
    \centering
    \begin{tabular}{c  c } 
     \hline\hline
     Parameter &  \\ 
     \hline\hline

    Nonlinear KAN architecture &   [3, 40, 40, 1]   \\ 
    $g_{nl}$   &             [5, 10, 15]               \\ 
    $k_{nl}$   &          3              \\ 
    Linear KAN architecture &    [3,  1]   \\ 
    $g_{l}$   &            1                \\ 
    $k_{l}$   &            1             \\ 
    
    High-fidelity learning rate &       0.005      \\ 
    High-fidelity iterations    &     100000       \\ 
    High-fidelity scales & [1, .8, .8] \\
    High-fidelity boundaries & [0, 20000, 40000] \\
    $N_{HF}$    &          100 time snapshots, 33153 points per snapshot     \\ 
    $w$ &  0  \\
    $n$ & 4 \\
    $\lambda_{\alpha}$ &  0.01\\
     \hline

                   Single-fidelity KAN architecture &   [3, 40, 40, 1] \\ 
    $g$   &        [5, 10, 15]                   \\ 
    $k$   &      3                  \\ 
    Single-fidelity learning rate &    0.01       \\ 
    Single-fidelity iterations    &  100000       \\ 
        Single-fidelity scales & [1, .8, .8] \\
    Single-fidelity boundaries & [0, 20000, 40000] \\
     \hline
    \end{tabular}
    \caption{Hyperparameters used for the results in~\cref{sec:Test7}.}
    \label{tab:params_Test7}
\end{table}

\end{document}